\documentclass[conference]{IEEEtran}
\IEEEoverridecommandlockouts

\usepackage{cite}
\usepackage{amsmath,amssymb,amsfonts}
\usepackage{algorithmic}
\usepackage{graphicx}
\usepackage{textcomp}
\usepackage{xcolor}
\usepackage{hyperref}
\usepackage{rotating}

\def\BibTeX{{\rm B\kern-.05em{\sc i\kern-.025em b}\kern-.08em
    T\kern-.1667em\lower.7ex\hbox{E}\kern-.125emX}}

\definecolor{Unlabeled}{RGB}{0, 0, 0}
\definecolor{Building}{RGB}{70,70,70}
\definecolor{Fence}{RGB}{190, 153, 153}
\definecolor{Other}{RGB}{72, 0, 90}
\definecolor{Pedestrian}{RGB}{220, 20, 60}
\definecolor{Pole}{RGB}{153, 153, 153}
\definecolor{Road line}{RGB}{157, 234, 50}
\definecolor{Road}{RGB}{128, 64, 128}
\definecolor{Sidewalk}{RGB}{244, 35, 232}
\definecolor{Vegetation}{RGB}{107, 142, 35}
\definecolor{Car}{RGB}{0, 0, 255}
\definecolor{Wall}{RGB}{102, 102, 156}
\definecolor{Traffic sign}{RGB}{220, 220, 0}

\begin{document}

\title{Semantic Label Reduction Techniques for Autonomous Driving}

\author{\IEEEauthorblockN{Qadeer Khan}
\IEEEauthorblockA{\textit{TUM and Artisense}}
\and
\IEEEauthorblockN{Torsten Sch\"on}
\IEEEauthorblockA{\textit{Audi Electronics Venture}}
\and
\IEEEauthorblockN{Patrick Wenzel}
\IEEEauthorblockA{\textit{TUM and Artisense}}
}

\maketitle

\begin{abstract}
Semantic segmentation maps can be used as input to models for maneuvering the controls of a car. However, not all labels may be necessary for making the control decision. One would expect that certain labels such as road lanes or sidewalks would be more critical in comparison with labels for vegetation or buildings which may not have a direct influence on the car's driving decision. In this appendix, we evaluate and quantify how sensitive and important the different semantic labels are for controlling the car. Labels that do not influence the driving decision are remapped to other classes, thereby simplifying the task by reducing to only labels critical for driving of the vehicle.
\end{abstract}

\section{Introduction}

\begin{figure}[ht]
  \centering
  \includegraphics[width=\linewidth]{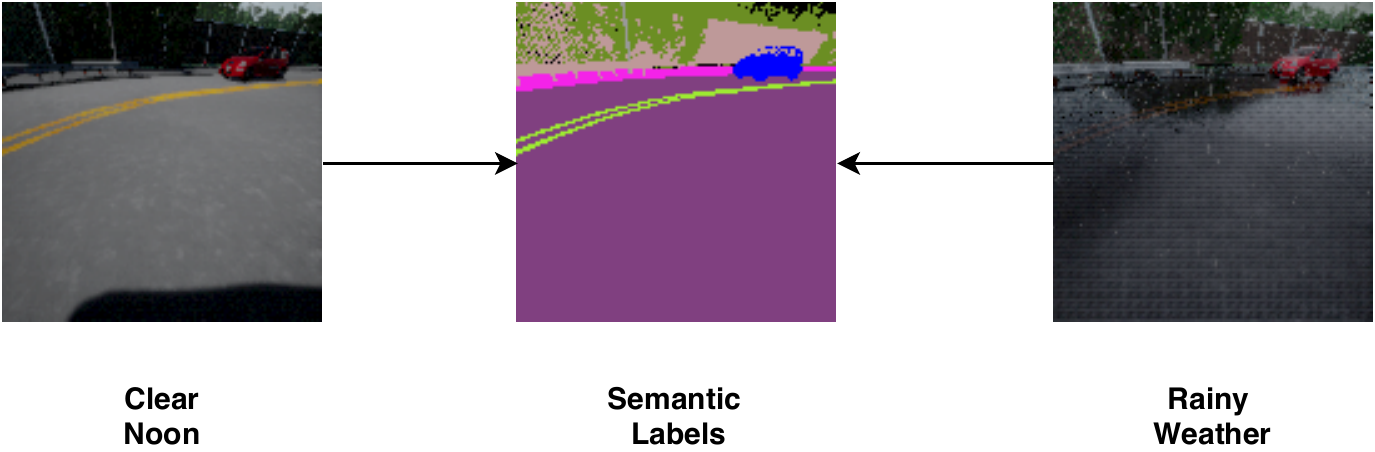}
  \caption{This figure shows an example of RGB images representing 2 different weather scenarios but with the same semantic representation.}
  \label{fig:whysegment}
\end{figure}

In the context of autonomous driving, semantic segmentation models are being widely used for the perception of the driving environment~\cite{MeyerIROS2018,WangarXiv2018} as well as for control of the ego-vehicle~\cite{MullerCoRL2018,WenzelCoRL2018}. Semantic maps offer several advantages over raw RGB data described below~\cite{WenzelCoRL2018}:

\begin{itemize}
\item Figure~\ref{fig:whysegment} shows how two weather conditions have different RGB inputs but the same semantic pixel labels. Hence, if the correct semantic representation is used as input to predict the correct steering commands, the model does not need to learn for each and every weather condition.
\item The semantic labels can precisely localize the pixels of important road landmarks such as traffic lights and signs. The status/information contained on these can then be read off to take appropriate planning and control decisions.
\item A high proportion of the pixels have the same label as its neighbours. This redundancy can be utilized to reduce the dimensionality of the semantic scene. Hence, the number of parameters required to train the control module can then also be reduced. \end{itemize}

Depending on the purpose of the semantic maps not all of its labels may be necessary. For example in predicting the steering angle of the ego-vehicle, certain labels such as road lines, sidewalks would be more important for the driving decision as opposed to labels for vegetation or buildings which do not have direct influence on the car's controls. In this paper, we evaluate how sensitive and important the different semantic labels are for controlling the driving behaviour. Labels that do not influence the driving decision are remapped to other classes. We use the following methods to identify the important of the semantic labels: 

\begin{itemize}
\item Grad-CAM (Section~\ref{section:grad_cam})
\item Semantic label removal (Section~\ref{section:label-removal})
\end{itemize}

Note that all the data collected and experiments performed are done using the CARLA simulator~\cite{Dosovitskiy2017} which provides semantic labels for 13 classes. These classes correspond to roads, sidewalks, road lines, fences, vehicles, pedestrian, other objects, vegetation, poles, traffic signs, walls, buildings, and a none class for objects that do not fall into any of the prior labels. We aim to control only the steering angle of the car while keeping the throttle fixed. The range of the steering varies between $-1$ and 1. The steering angle in degrees corresponding to these values depends on the vehicle being used. In our case, the default vehicle is used for which for the the maximum steering angle is 70$^\circ$. Videos referenced to in the subsequent sections could be found at the following link:~\url{https://www.youtube.com/playlist?list=PLKWxSGEZd0AcvsC2N5trDhPbbeLdGLfi3}.

\section{Related Work}\label{section:relwork_sssfusion}

\noindent{\textbf{Semantic Sensitivity.}} The authors of~\cite{Chen2017a} instituted a hierarchical structure for segregating the various semantic classes based on their relative importance by assigning them weights proportionate to their significance. Classes with differing importance were placed on the different levels of the hierarchy. A special "Importance Aware Loss" was introduced that stressed the need for correctly segmenting the more important semantic classes in comparison to less important ones.

\section{Grad-CAM}\label{section:grad_cam}

\begin{figure}[ht]
  \centering
  \includegraphics[width=\linewidth]{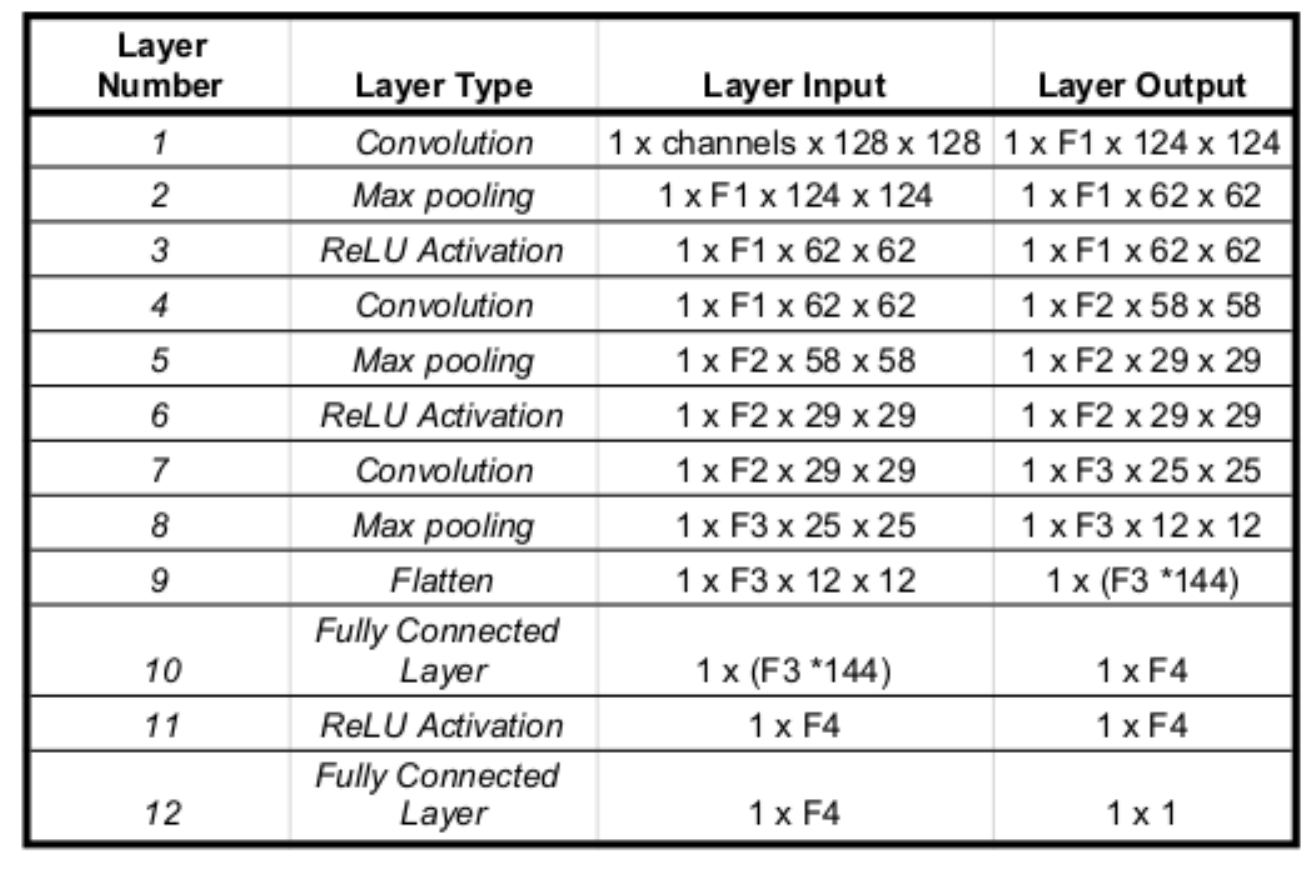}
  \caption{The architecture for training a model controlling the steering command of a car using RGB/segmentation images. In the first layer \textit{channels} = 3 (for RGB input) or 13 (for segmentation input). $F_1$, $F_2$, and $F_3$ are the number of filters of the convolution operations represented by layers 1, 4, and 7, respectively. $F_4$ are the number of neurons in the fully connected Layer 10. All convolutional layers have a kernel size of 5 and a stride of 1. All maxpooling layers have a kernel size of 2 and stride of 2.}
  \label{fig:SemSensArchi}
\end{figure}

Gradient-weighted class activation maps (Grad-CAM) is a technique providing visual explanation on how a model classifies images~\cite{ChattopadhyayWACV2018,SelvarajuICCV2017}. This is done by tracking the flow of gradients, to localize areas in the original input image that are important to the model for making the correct classification. Our task of predicting the steering command is that of regression rather than classification. Nevertheless, by using the same Grad-CAM technique of tracking the flow of gradients, we can ascertain regions in the input RGB image and ultimately the corresponding semantic labels that are fundamental for the model to produce the correct steering command. We train a simple end-to-end model, whose architecture is described in Figure~\ref{fig:SemSensArchi}. The model has the following parameters: \textit{channels} = 3, $F_1 = 3$, $F_2 = 3$, $F_3 = 13$, $F_4 = 4$.

\begin{figure}[ht]
  \centering
  \includegraphics[width=\linewidth]{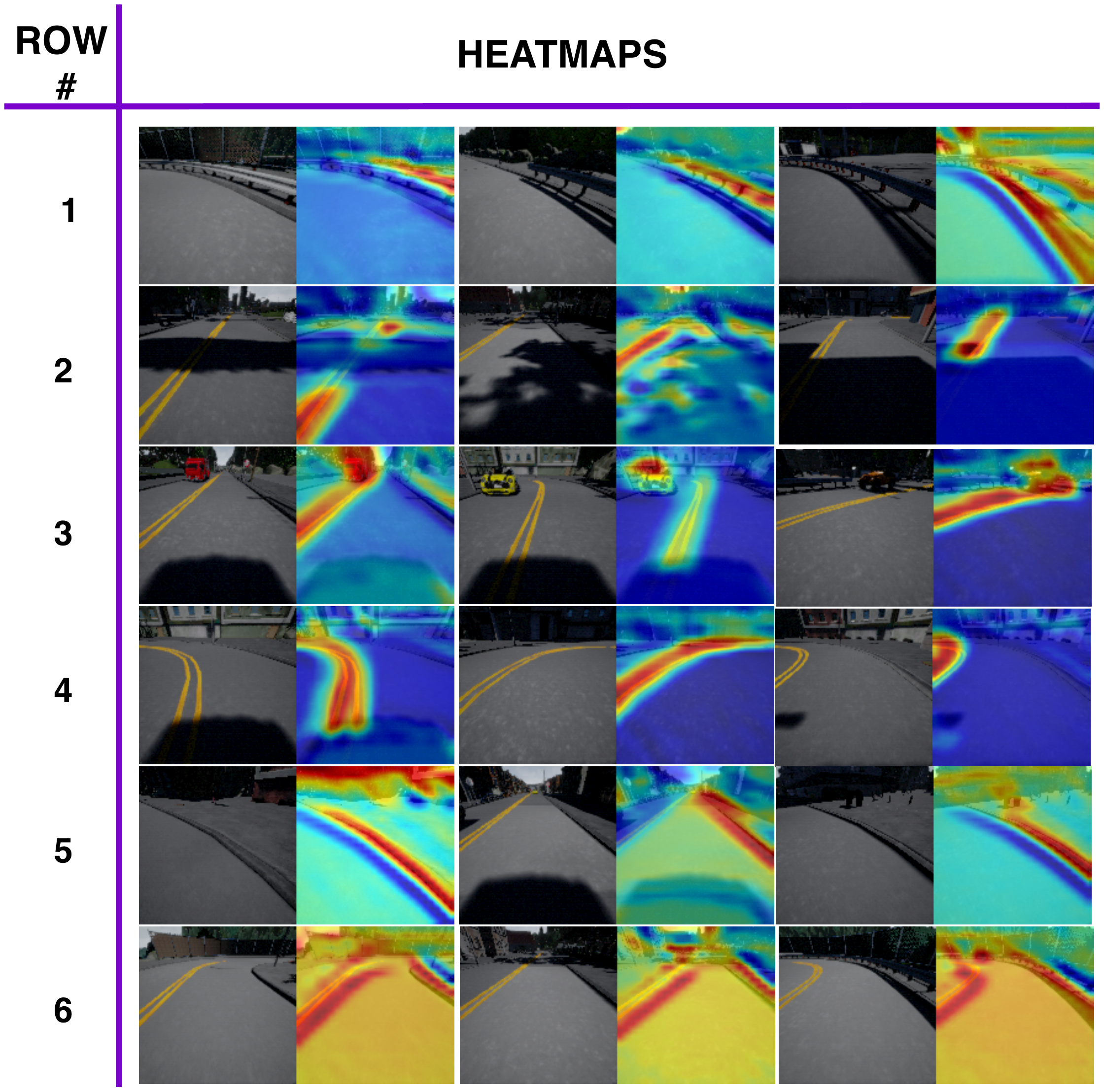}
  \caption{Some sample images and affiliated heat maps. While taking the appropriate steering decision, the model seems to be looking into regions for fences (row 1), road lines (row 2 and 4), vehicles (row 3), sidewalk boundary (row 5), and the road (row 6) itself. In row 2, the model is skipping shadows and only focusing on the visible road lines for decision making. Dark red represents regions of high importance, while light blue color represents portions of low relevance for the model for decision making.}
  \label{fig:GradCamExamples}
\end{figure}

We track the backward gradient flow from layer 7 of the model right up till the input image, to hone in on regions that are of interest for driving. Figure~\ref{fig:GradCamExamples} shows these regions overlaid on the original image as a heat map, where the color of the heat map corresponds to the intensity of relevance. Dark red color represents high importance, whereas light blue corresponds to regions of low relevance. It can be observed that the important regions for driving decision correspond to those with semantic labels of fences (row 1), road lines (row 2 and 4), vehicles (row 3), sidewalks boundary with the road (row 5), and the road (row 6) itself. It is also interesting to note from sample images in row 2 that the model seems to be skipping the shadows and only basing its decisions on the road lines. For certain samples of continuous driving sessions, video1 demonstrates regions in the input RGB image which the model uses for decision making. 

\section{Semantic Label Removal Techniques}\label{section:label-removal}

In this approach, we train an end-to-end model with the semantic labels as input. The model input has 13 channels, corresponding to the 13 semantic labels. The architecture of this model is described in Figure~\ref{fig:SemSensArchi} with the following parameters, \textit{channels} = 13, $F_1 = 5$, $F_2 = 4$, $F_3 = 14$, $F_4 = 5$.

\begin{figure}[ht]
  \centering
  \includegraphics[width=\linewidth]{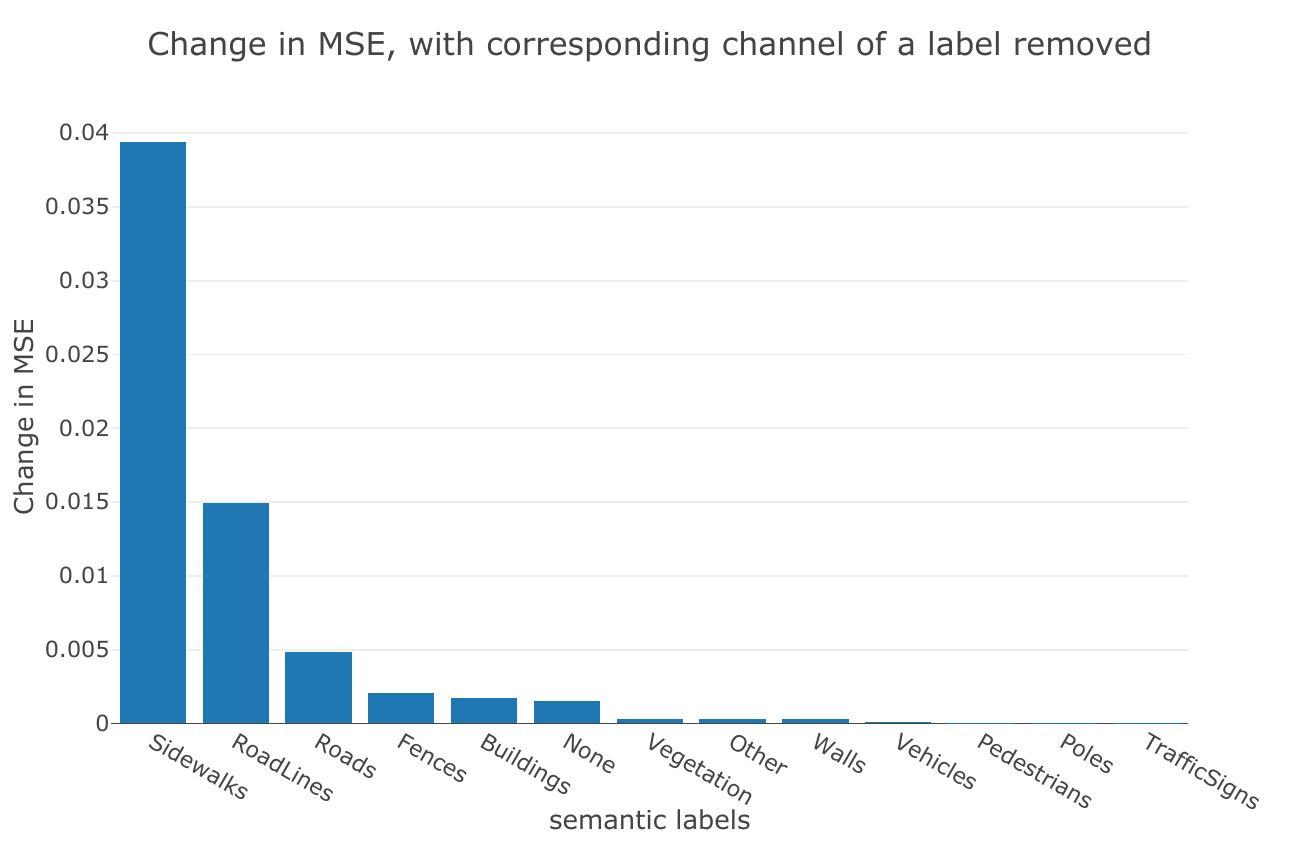}
  \caption{Increase in error due to removal of each of the 13 semantic labels by zeroing out the channel corresponding to that label. The bar plot is arranged in descending order of error. The same order also provides an indication of the relative importance of that label towards making a driving decision.  }
  \label{fig:sem_Sensitvity}
\end{figure}

Once this model is trained, we evaluate the sensitivity of each semantic label by feeding zeros to its corresponding channel and recording the change in error. The error is the mean squared error (MSE) between the steering angle predicted by the model and the ground truth. Zeroing out a channel effectively eliminates that label from the input. Removal of semantic labels which are critical towards making the driving decision would result in an increased error. Figure~\ref{fig:sem_Sensitvity} shows the increase in error by removal of each of the semantic labels.

The bar plot is arranged in descending order of error and hence of label importance. From this method we observe that the error increases dramatically if we are to remove channels corresponding to sidewalks and road lines. Next are the roads and fence labels in order of importance. This method also seems to reaffirm our observation from the Grad-Cam approach in Section \ref{section:grad_cam} that road lines and sidewalks are of utmost importance for the control model to execute the correct steering command.

\section{Label Remapping}\label{section:label_remapping}

\begin{figure}[ht]
  \centering
  \includegraphics[width=\linewidth]{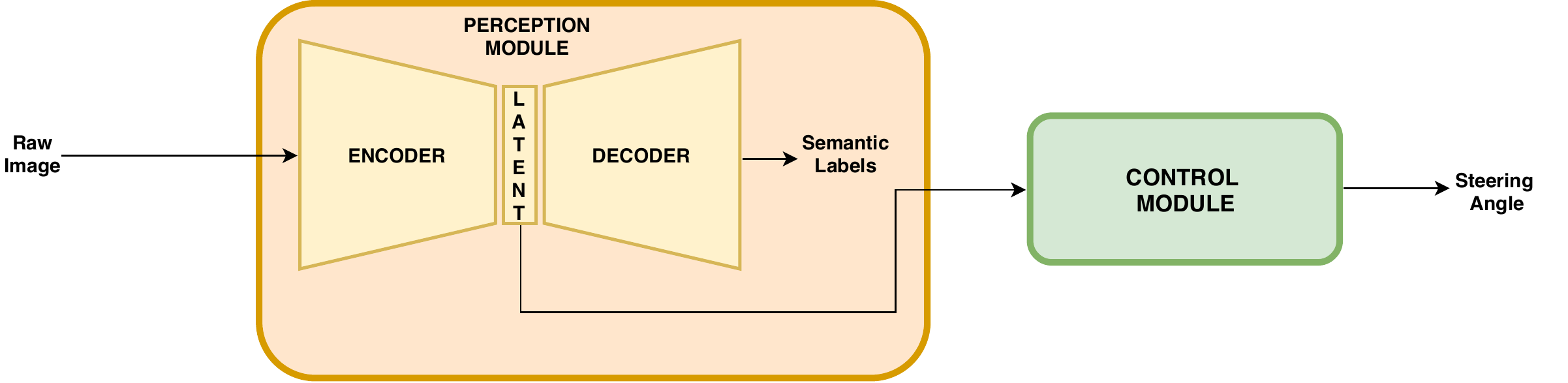}
  \caption{The perception module is trained as an encoder-decoder architecture, without any skip connections. The encoder sub-module first embeds the raw image into a lower dimensional latent vector. The decoder sub-module reconstructs the semantic scene from this latent vector. If the low-dimensional latent vector contains all the necessary information to reconstruct the semantic scene to a reasonable degree of accuracy, then we directly feed it as an input to the control module instead of the semantic labels. The architecture is same as that used by~\cite{WenzelCoRL2018}.}
  \label{fig:segcodercontrol}
\end{figure}

To predict the steering angle of the car, we train the model with a modular approach instead of an end-to-end learning approach, for reasons given in~\cite{WenzelCoRL2018}. This approach is shown in Figure~\ref{fig:segcodercontrol}. The purpose of the perception module is to use the images captured by an RGB camera to extract semantic features of the scene. These extracted features are then fed to the control module which aims to produce the correct steering command.

\begin{figure}[ht]
  \centering
  \includegraphics[width=\linewidth]{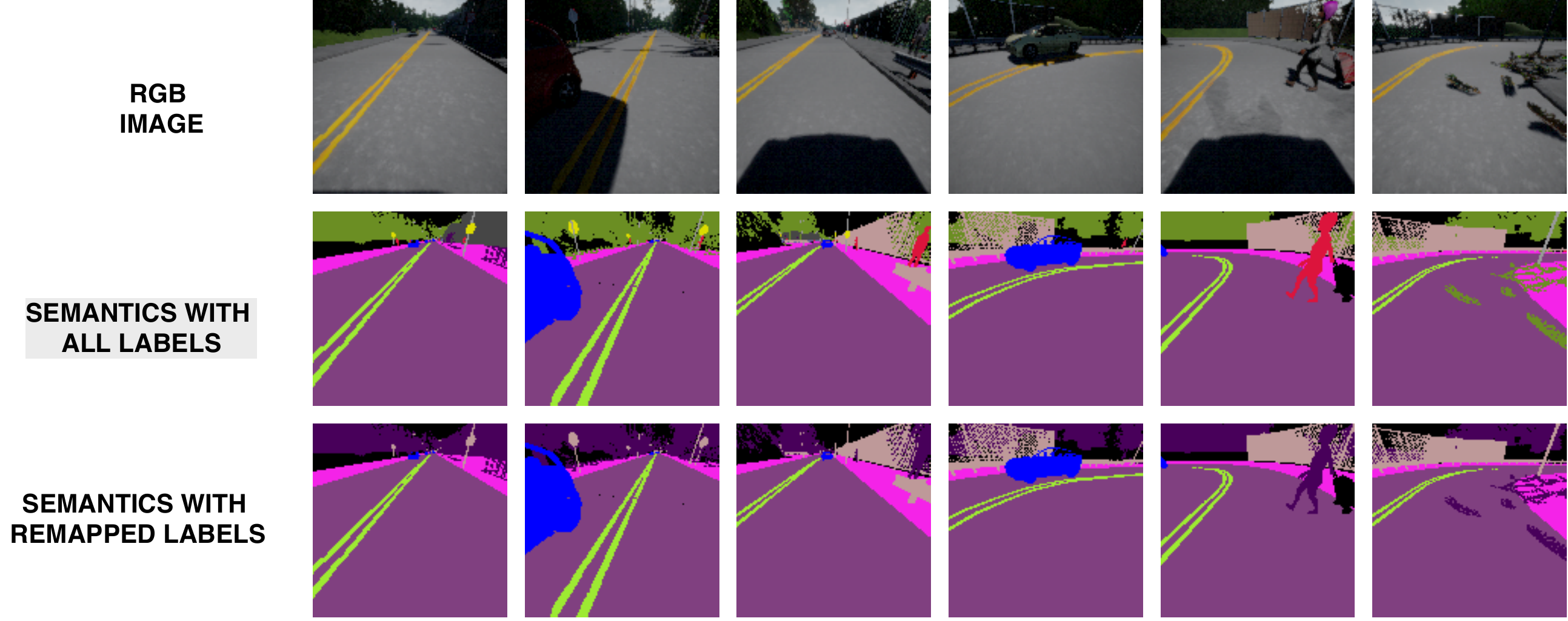}
  \caption{This figure shows 5 examples of how remapping influences the visual appearance of the semantic map. The 1\textsuperscript{st} row contains the raw RGB image, the 2\textsuperscript{nd} row are the corresponding semantic maps with all the 13 labels. The 3\textsuperscript{rd} row are the semantics with labels remapped in accordance with Table \ref{tab:remapping}. Note, from the images in the 3\textsuperscript{rd} row that those containing a \emph{pedestrian} (5\textsuperscript{th} column), \emph{vegetation} (6\textsuperscript{th} column)  are remapped to the \emph{other} class. Similarly, labels for \emph{poles} are remapped to the \emph{fence} class (1\textsuperscript{st} and 2\textsuperscript{nd} columns).}
  \label{fig:remappedExamples}
\end{figure}

However, as demonstrated earlier, not all the semantic labels may be important for the driving strategy. Reconciling the conclusions from the 2 methods presented in Section~\ref{section:grad_cam} and Section~\ref{section:label-removal}, we train a new perception module by remapping the semantic labels as described in Table~\ref{tab:remapping}. Figure~\ref{fig:remappedExamples} shows some visualizations of the semantic maps resulting from this remapping.

\begin{table}[ht]
    \centering
    \caption{The first and third rows of the table enlists the 13 semantic labels. The second and forth rows shows which labels they are remapped to. Note that the color of text of each label corresponds to the color of their semantic representation displayed in Figure~\ref{fig:remappedExamples}.}
    \resizebox{\linewidth}{!}{  
    \begin{tabular}{|l|ccccccc|}
        \hline
        \textbf{Label} & \textcolor{Road}{Roads} & \textcolor{Sidewalk}{Sidewalks} & \textcolor{Road line}{RoadLines} & \textcolor{Fence}{Fences} & \textcolor{Car}{Vehicles} & \textcolor{Pedestrian}{Pedestrian} & \textcolor{Other}{Other}\\
        \hline
        \textbf{Mapped to}  & \textcolor{Road}{Roads} & \textcolor{Sidewalk}{Sidewalks} & \textcolor{Road line}{RoadLines} & \textcolor{Fence}{Fences} & \textcolor{Car}{Vehicles} & \textcolor{Other}{Other} & \textcolor{Other}{Other}  \\

        \hline
        \textbf{Label} &   \textcolor{Vegetation}{Vegetation} & \textcolor{Pole}{Poles} & \textcolor{Traffic sign}{Traffic Sign} & \textcolor{Wall}{Wall} & \textcolor{Building}{Building} & \textcolor{Unlabeled}{None} & \\
        
        \hline

        \textbf{Mapped to}  & \textcolor{Other}{Other} & \textcolor{Fence}{Fences} & \textcolor{Fence}{Fences} & \textcolor{Other}{Other} & \textcolor{Other}{Other} & \textcolor{Unlabeled}{None} & \\
        \hline
        
    \end{tabular}

    \label{tab:remapping}
    }
\end{table}

Table~\ref{tab:remapping_errors} gives a comparison of the training, validation, and test performance of the 2 models trained on the normal and the remapped segmentation labels. It can be observed that the remapping of labels does not cause any degradation in performance. In fact, a comparison of video2 (perception trained with all labels) and video3 (perception trained with remapped labels) shows that the remapped perception module produces a more stable segmentation reconstruction. 

\begin{table}[ht]
    \centering
    \caption{The first row is the error for the control model whose perception module is trained with all labels, while the second row is for the perception module trained with the new remapped labels. The values in the table are the mean squared error (MSE) between the actual and the steering command predicted by the models represented in the order of $10^{-3}$.}
    \resizebox{\linewidth}{!}{  
    \begin{tabular}{l|ccc}
        \hline
        \textbf{Segmentation Model} & Training Error & Validation Error & Test Error \\
        \hline
        \textbf{With all labels}  &  5.68 & 9.59 &  9.15 \\
        \textbf{With remapped labels}  & 4.64 & 9.20 & 9.11  \\
        \hline
    \end{tabular}
    \label{tab:remapping_errors}
    }
\end{table}

\subsection{Discussion}
The manner in which the remapping is to be done or how the sensitivity is calculated is arguable. For e.g. instead of just zeroing out the label for road lines it might have been worthwhile to see what would happen if we  camouflage it into the road. This could be done by assigning the road line label to also be a road in the road channel. There could be a multitude of such possibilities to be considered, which may grow exponentially as we refine the segmentation map by increasing the number of semantic labels. Finding the important semantic labels is in itself a separate research topic. As explained in Section~\ref{section:relwork_sssfusion}, the authors of~\cite{Chen2017a} have tried to address this problem in the context of autonomous driving.

\section{Conclusion}
From the results of the experiments, we observed that training a perception module by remapping the less important semantic labels to other classes, did not lead to any degradation in model performance. The car was still able to be controlled under this approach. This has a positive repercussion in that the semantic classes to be labeled can be reduced thereby possibly curtailing the human effort.

\bibliographystyle{IEEEtran}
\bibliography{main}

\end{document}